\pdfoutput=1
\documentclass[11pt]{article}

\usepackage{acl}

\usepackage{times}
\usepackage{latexsym}

\usepackage[T1]{fontenc}

\usepackage[utf8]{inputenc}

\usepackage{microtype}

\usepackage{comment}
\usepackage{amsmath}
\usepackage{amssymb}
\usepackage{bm}

\usepackage{dsbda}

\usepackage{booktabs}
\usepackage{multirow}
\usepackage{xspace}
\usepackage{graphicx}

\newcommand{\mlp}{WideMLP\xspace}

\title{Bag-of-Words vs.\@ Graph vs.\@ Sequence in Text Classification: Questioning the Necessity of Text-Graphs and the Surprising Strength of a Wide MLP}

\author{Lukas Galke \\
  University of Kiel, Germany\\
  MPI for Psycholinguistics, the Netherlands\\
  \texttt{lukas.galke@mpi.nl} \\\And
  Ansgar Scherp \\
  University of Ulm, Germany \\
  \texttt{ansgar.scherp@uni-ulm.de} \\}

\begin{document}

\maketitle

\begin{abstract}
Graph neural networks have triggered a resurgence of graph-based text classification methods, defining today's state of the art. 
We show that a wide multi-layer perceptron (MLP) using a Bag-of-Words (BoW) outperforms the recent graph-based models TextGCN and Hete\-GCN in an inductive text classification setting and is comparable with HyperGAT. 
Moreover, we fine-tune a sequence-based BERT and a lightweight DistilBERT model, which both outperform all state-of-the-art models. 
These results question the importance of synthetic graphs used in modern text classifiers.
In terms of efficiency, DistilBERT is still twice as large as our BoW-based wide MLP, while
graph-based models like TextGCN require setting up an $\mathcal{O}(N^2)$ graph, where $N$ is the vocabulary plus corpus size. 
Finally, since Transformers need to compute $\mathcal{O}(L^2)$ attention weights with sequence length $L$, the MLP models show higher training and inference speeds on datasets with long sequences. 
\end{abstract}

\section{Introduction}

Text categorization is the task of assigning topical categories to text units such as documents, social media postings, or news articles.
Research on text categorization is a very active field as just the sheer amount of new methods in recent surveys shows ~\cite{DBLP:journals/corr/abs-2107-03158,DBLP:journals/corr/abs-2008-00364,DBLP:journals/wias/ZhouGLVTBBK20, DBLP:journals/information/KowsariMHMBB19,DBLP:journals/air/Kadhim19}.

There are approaches based on a Bag of Words (BoW) that perform text categorization purely on the basis of a multiset of tokens.
Among them are 
Deep Averaging Networks (DAN)~\cite{DBLP:conf/acl/IyyerMBD15}, a deep Multi-Layer Perceptron (MLP) model with $n$ layers that relies on averaging the BoW, Simple Word Embedding Models (SWEM)~\cite{DBLP:conf/acl/HenaoLCSSWWMZ18} that explores different pooling strategies for pretrained word embeddings, and fastText~\cite{DBLP:journals/tacl/BojanowskiGJM17}, which uses a linear layer on top of pretrained word embeddings.
These models count the occurrence of all tokens in the input sequence, while disregarding word position and order, and then rely on word embeddings and fully connected feedforward layer(s).
We call these \newterm{BoW-based models}. 

Among the most popular recent methods for text categorization are graph-based models such as TextGCN~\cite{DBLP:conf/aaai/YaoM019} that first induce a synthetic word-document co-occurence graph over the corpus and subsequently apply a graph neural network (GNN) to perform the classification task.
Besides TextGCN, there are follow-up works like
HeteGCN~\cite{DBLP:conf/wsdm/RageshSIBL21}, TensorGCN~\cite{DBLP:conf/aaai/LiuYZWL20}, and HyperGAT~\cite{DBLP:conf/emnlp/DingWLLL20}, which we collectively call \newterm{graph-based models}.

Finally, there is the well-known Transformer~\cite{DBLP:conf/nips/VaswaniSPUJGKP17} universe with models such as BERT~\cite{DBLP:conf/naacl/DevlinCLT19} and its size-reduced variants such as DistilBERT~\cite{distilbert}.
Here, the input is a (fixed-length) sequence of tokens, which is then fed into multiple layers of self-attention.
Lightweight versions such as DistilBERT and others~\cite{DBLP:journals/corr/abs-2009-06732,DBLP:journals/corr/abs-2103-14636} use less parameters but operate on the same type of input.
Together with recurrent models such as LSTMs, we call these \newterm{sequence-based models}.

In this paper, we hypothesize that text categorization can be very well conducted by simple but effective BoW-based models.
We investigate this research question in three steps:
First, we conduct an in-depth analysis of the literature.
We review the key research in the field of text categorization.
From this analysis, we derive the different families of methods, the established benchmark datasets, and identify the top performing methods.
We decide for which models we report numbers from the literature and which models we run on our own. 
Overall, we compare 16 different methods from the families of BoW-based models (8 methods), sequence-based models (3 methods), and graph-based models (5 methods).
We run our own experiments for 7 of these methods on 5 text categorization datasets, while we report the results from the literature for the remaining methods.

The result is surprising: 
Our own BoW-based MLP, called the \mlp, with only one wide hidden layer, outperforms many of the recent graph-based models for inductive text categorization~\cite{DBLP:conf/aaai/YaoM019,DBLP:conf/aaai/LiuYZWL20,DBLP:conf/wsdm/RageshSIBL21}. 
Moreover, we did not find any reported scores for BERT-based methods from the sequence-based family. 
Thus, we fine-tuned our own  BERT~\cite{DBLP:conf/naacl/DevlinCLT19} and DistilBERT~\cite{distilbert}.
These models set a new state of the art.
On a meta-level, our study shows that MLPs have largely been ignored as competitor methods in experiments.
It seems as if MLPs have been forgotten as baseline in the literature, which instead is focusing mostly on other advanced Deep Learning architectures.
Considering strong baselines is, however, an important means to argue about \textit{true} scientific advancement~\cite{DBLP:conf/acl/HenaoLCSSWWMZ18,DBLP:conf/recsys/DacremaCJ19}.
Simple models are also often preferred in industry due to lower operational and maintenance costs.

Below, we introduce our methodology and results from the literature study.
Subsequently, we introduce the families of models in \Secref{sec:methods}. Thereafter, we describe the experimental procedure in \Secref{sec:apparatus}.
We present the results of our experiments in \Secref{sec:results} and discuss our findings in \Secref{sec:discussion}, before we conclude.

\section{Literature on Text Categorization}\label{sec:sota}

\paragraph{Methodology}
In a first step, we have analyzed recent surveys on text categorization and comparison studies~\cite{DBLP:journals/csur/MinaeeKCNCG21,DBLP:journals/corr/abs-2107-03158,DBLP:journals/corr/abs-2008-00364,DBLP:journals/wias/ZhouGLVTBBK20,DBLP:journals/information/KowsariMHMBB19,DBLP:journals/air/Kadhim19,DBLP:conf/kcap/GalkeMSBS17,DBLP:conf/sigir/ZhangWYWZ16}.
These cover the range from shallow to deep classification models.
Second, we have screened for literature in key NLP and AI venues.
Finally, we have complemented our search by checking results and papers on \href{https://paperswithcode.com/task/text-classification}{paperswithcode.com}.
On the basis of this input, we have determined three families of methods and benchmark datasets (see Table~\ref{tab:datasets}).
We focus our analysis on identifying models per family showing strong performance and select the methods to include in our study.
For all models, we have verified that the same train-test split is used.
We check whether modified versions of the datasets have been used (\eg fewer classes), to avoid bias and wrongfully giving advantages.

\paragraph{BoW-based Models}
Classical machine learning models that operate on a BoW-based input are extensively discussed in two surveys~\cite{DBLP:journals/information/KowsariMHMBB19,DBLP:journals/air/Kadhim19} and other comparison studies~\cite{DBLP:conf/kcap/GalkeMSBS17}.
\citet{DBLP:conf/acl/IyyerMBD15} proposed DAN, which combine word embeddings and deep feedforward networks.
It is an MLP with 1-6 hidden layers, non-linear activation, dropout, and AdaGrad as optimization method.
The results suggest to use pretrained embeddings such as GloVe~\cite{DBLP:conf/emnlp/PenningtonSM14} over a randomly initialized neural bag of-words~\cite{DBLP:conf/acl/KalchbrennerGB14} as input.
In fastText~\cite{DBLP:journals/tacl/BojanowskiGJM17,DBLP:conf/eacl/GraveMJB17} a linear layer on top of pretrained embeddings is used for classification.
Furthermore, \citet{DBLP:conf/acl/HenaoLCSSWWMZ18} explore embedding pooling variants and find that SWEM can rival  approaches based on recurrent (RNN) and convolutional neural networks (CNN).
We consider fastText, SWEM, and a DAN-like deeper MLP in our comparison.

Note that those approaches that rely on logistic regression on top of pretrained word embeddings, \eg fastText, share a similar architecture as an MLP with one hidden layer. 
However, the standard training protocol involves pretraining the word embedding on large amounts of unlabeled text and then freezing the word embeddings while training the logistic regression~\cite{DBLP:conf/nips/MikolovSCCD13}.

\textbf{Graph-based Models} 
Using graphs induced from text for the task of text categorization has a long history in the community.
An early work is the term co-occurrence graph of the KeyGraph algorithm~\cite{DBLP:conf/adl/OhsawaBY98}.
The graph is split into segments, representing the key concepts in the document.
Co-occurence graphs have also been used for automatic keyword extraction such as in RAKE~\cite{Rose2010} and can be also used for classification~\cite{DBLP:conf/emnlp/ZhangDXLZ21}.

Modern approaches exploit this idea in combination with 
graph neural networks (GNN)~\cite{book:hamilton:grl}.
Examples of GNN-based methods operating on a word-document co-occurence graph are TextGCN~\cite{DBLP:conf/aaai/YaoM019} and its successor TensorGCN~\cite{DBLP:conf/aaai/LiuYZWL20} as well as  Hete\-GCN~\cite{DBLP:conf/wsdm/RageshSIBL21},
HyperGAT~\cite{DBLP:conf/emnlp/DingWLLL20}, and
DADGNN~\cite{DBLP:conf/aaai/LiuYZWL20}.
We briefly discuss these models:
In TextGCN, the authors set up a graph based on word-word connections given by window-based pointwise mutual information and word-document TF-IDF scores.
They use a one-hot encoding as node features and apply a two-layer graph convolutional network~\cite{DBLP:conf/iclr/KipfW17} on the graph to carry out the node classification task.
HeteGCN combines ideas from Predictive Text Embedding~\citep{DBLP:conf/kdd/TangQM15} and TextGCN and split the adjacency matrix into its word-document and word-word sub-matrices and fuse the different layers' representations when required.
TensorGCN uses multiple ways of converting text data into graph data including a semantic graph created with an LSTM, a syntactic graph created by dependency parsing, and a sequential graph based on word co-occurrence. 
HyperGAT extended the idea of text-induced graphs for text classification to hypergraphs.
The model uses graph attention and two kinds of hyperedges.
Sequential hyperedges represent the relation between sentences and their words.
Semantic hyperedges for word-word connections are derived from topic models~\cite{DBLP:journals/jmlr/BleiNJ03}.
Finally, DADGNN is a graph-based approach that uses attention diffusion and decoupling techniques to tackle oversmoothing of the GNN and to be able to stack more layers.

In TextGCN's original transductive formulation, the entire graph including the test set needs to be known for training. This may be prohibitive in practical applications as each batch of new documents would require retraining the model.
When these methods are adapted for inductive learning, where the test set is unseen, they achieve notably lower scores~\cite{DBLP:conf/wsdm/RageshSIBL21}.
GNNs for text classification use corpus statistics, \eg pointwise mutual information (PMI), to connect related words in a graph~\cite{DBLP:conf/aaai/YaoM019}. When these were omitted, the GNNs would collapse to bag-of-words MLPs. Thus, GNNs have access to more information than BoW-MLPs. GloVe~\cite{DBLP:conf/emnlp/PenningtonSM14} also captures PMI corpus statistics, which is why we include an MLP on GloVe input representations.

\paragraph{Sequence models: RNN and CNN}
Recurrent neural networks (RNN) are a natural choice for any NLP task. 
However, it turned out to be challenging to find numbers reported on text categorization in the literature that can be used as references.
The bidirectional LSTM with two-dimensional max pooling BLSTM-2DCNN~\cite{DBLP:conf/coling/ZhouQZXBX16} has been applied on a stripped-down to 4 classes version of the 20ng dataset. 
Thus, the high score of $96.5$ reported for 4ng cannot be compared with papers applied on the full 20ng dataset.
Also Text\-RCNN~\cite{DBLP:conf/aaai/LaiXLZ15}, a model combining recurrence and convolution 
uses only the 4 major categories in the 20ng dataset. 
The results of Text\-RCNN are identical with BLSTM-2DCNN.
For the MR dataset, BLSTM-2DCNN provides no information on the specific split of the dataset.
RNN-Capsule~\cite{DBLP:conf/www/WangSH0Z18} is a sentiment analysis method reaching
an accuracy of $83.8$ on the MR dataset, but with a different train-test split. 
\citet{DBLP:journals/corr/abs-2006-15795} combine a 2D-CNN with bidirectional RNN.
Another work applying a combination of a convolutional layer and an LSTM layer is by \citet{DBLP:conf/ijcnn/WangLCCW19}. 
The authors experiment with five English and two Chinese datasets, which are not in the set of representative datasets we identified.
The authors report that their approach outperforms existing models like fastText on two of the five English datasets and both Chinese datasets. 

\paragraph{Sequence models: Transformers}
Surprisingly, only few works consider Transformer models for text categorization.
A recent work shows that BERT outperforms classic TF-IDF BoW approaches on English, Chinese, and Portuguese text classification datasets~\citep{gonzales-2020-comparing-bert}.
We have not found any results of transformer-based models reported on those text categorization datasets that are commonly used in the graph-based approaches.

Therefore, we fine-tune BERT~\cite{DBLP:conf/naacl/DevlinCLT19} and DistilBERT~\cite{distilbert} on those datasets ourselves. BERT is a large pretrained language model on the basis of Transformers.
DistilBERT~\citep{distilbert} is a distilled version of BERT with 40\% reduced parameters while retaining 97\% of BERT's language understanding capabilities. 
TinyBERT~\citep{tinybert} and MobileBERT~\citep{sun2020mobilebert} would be similarly suitable alternatives, among others. 
We chose DistilBERT because it can be fine-tuned independently from the BERT teacher. 
Its inference times are 60\% faster than BERT, which makes it more likely to be reusable by labs with limited resources.

\paragraph{Summary}
From our literature survey, we see that all recent methods are based on graphs. 
BoW-based methods are hardly found in experiments, while, likewise surprisingly, Transformer-based sequence models are extremely scarce in the literature on topical text categorization.
The recent surveys on text categorization include both classical and Deep Learning models, but none considered a simple MLP except for the inclusion of DAN~\cite{DBLP:conf/acl/IyyerMBD15} in \citet{DBLP:journals/corr/abs-2008-00364}. 

\begin{table*}[ht]
  \centering
  \small
  \begin{tabular}{lcccc}
    \toprule
    \textbf{Model} & \textbf{Synthetic Graph} & \textbf{Position-Aware} & \textbf{Arbitrary Length} & \textbf{Inductive} \\
    \midrule
    Bag-of-Words & No & No  &   Yes & Yes\\
    Graph: TextGCN & Yes & No & Yes & No\\
    Graph: TensorGCN & Yes & Yes & Yes & No\\ 
    Graph: HeteGCN/HyperGAT & Yes & No & Yes & Yes\\
    Sequence: RNN/CNN & No & Yes  & Yes & Yes\\
    Sequence: BERT/DistilBERT & No & Yes  & No & Yes\\
    \bottomrule
  \end{tabular}
  \caption{Properties of text categorization approaches. Graph-based models that rely on having access to unlabeled test documents such as TextGCN and TensorGCN are not capable of inductive learning without modifications.}
  \label{tab:properties}
\end{table*}

\section{Models for Text Categorization}\label{sec:methods}

We formally introduce the three families of models for text categorization, namely the BoW-based, graph-based, and sequence-based models.
Table~\ref{tab:properties} summarizes the key properties of the approaches: whether they require a synthetic graph, whether word position is reflected in the model, whether the model can deal with arbitrary length text, and whether the model is capable of inductive learning.

\subsection{BoW-Based Text Categorization}
Under pure BoW-based text categorization, we denote approaches that are not order-aware and operate only on the multiset of words from the input document. 
Given paired training examples $(\vx,y) \in \train$, each consisting of a bag-of-words $\vx \in \sR^{n_\mathrm{vocab}}$ and a class label  
$y \in \sY$, the goal is to learn a generalizable function $\hat\vy = f_\theta^{(\mathrm{BoW})}(\vx)$ with parameters $\theta$ such that $\argmax(\hat\vy)$ preferably equals the true label $y$ for input $\vx$.

As BoW-based model, we consider a one hidden layer \mlp (\ie two layers in total). 
We experiment with pure BoW, TF-IDF weighted, and averaged GloVe input representations.
We also use a two hidden layers \mlp-2.
We list the numbers for fastText, SWEM, and logistic regression from \citet{DBLP:conf/emnlp/DingWLLL20} in our comparison.

\subsection{Graph-Based Text Categorization}

Graph-based text categorization approaches first set up a \emph{synthetic} graph on the basis of the text corpus $\train$ in the form of an adjacency matrix $\hat\mA := \operatorname{make-graph}(\train)$.
For instance, in TextGCN the graph is set up in two parts: word-word connections are modeled by pointwise mutual information and word-document edges resemble that the word occurs in the document.
Then, a parameterized function $f_\theta^{(\mathrm{graph})}(\mX, \hat\mA)$ is learned that uses the graph as input, where $\mX$ are the node features. 
The graph is composed of word and document nodes, each receiving its own embedding (by setting $\mX =\mI$).
In inductive learning, however, there is no embedding of the test documents.
Note that the graph-based approaches from the current literature such as TextGCN also disregard word order, similar to the BoW-based models described above.
A detailed discussion of the connection between TextGCN and MLP is provided in Appendix~\ref{app:textgcn-vs-mlp}.

We consider top performing graph-based models from the literature, namely TextGCN along with its successors HeteGCN, TensorGCN, HyperGAT, DADGNN, as well as simplified GCN (SGC)~\cite{DBLP:conf/icml/WuSZFYW19}. 
We do not run our own experiments for the graph-based models but rely on the original work and extensive studies by \citet{DBLP:conf/emnlp/DingWLLL20} and \citet{DBLP:conf/wsdm/RageshSIBL21}. 

\subsection{Sequence-Based Text Categorization}

We consider RNNs, LSTMs, and Transformers as sequence-based models. 
These models are aware of the order of the words in the input text in the sense that they are able to exploit word order information.
Thus, the key difference to the BoW-based and graph-based families is that the word order is reflected by sequence-based model. The model signature is
$\hat{y} = f_\theta^{( \mathrm{sequence})} ( \langle x_1, x_2, \ldots, x_k \rangle )$,
where $k$ is the (maximum) sequence length.
Word position is modeled by a dedicated positional encoding. For instance, in BERT each position is associated with an embedding vector that is added to the word embedding at input level.

For the sequence-based models, we run our own experiments with BERT and DistilBERT, while reporting the scores of a pretrained LSTM from \citet{DBLP:conf/emnlp/DingWLLL20} for comparison.

\section{Experimental Apparatus}\label{sec:apparatus}

\subsection{Datasets}
We use the same datasets and train-test split as in TextGCN~\cite{DBLP:conf/aaai/YaoM019}.
Those datasets are 20ng, R8, R52, ohsumed, and MR.
  Twenty Newsgroups (20ng)\footnote{\url{http://qwone.com/~jason/20Newsgroups/}} (bydate version) contains long posts  categorized into 20 newsgroups.
  The mean sequence length is 551 words with a standard deviation (SD) of 2,047.
  R8 and R52 are subsets of the Reuters~21578 news dataset with 8 and 52 classes, respectively.  The mean sequence length and SD is $119\pm128$ words for R8, and $126 \pm 133$ words for R52.
  Ohsumed\footnote{\url{http://disi.unitn.it/moschitti/corpora.htm}} is a corpus of medical abstracts from the MEDLINE database that are categorized into diseases (one per abstract).
  The mean sequence length is $285 \pm 123$ words.
  Movie Reviews (MR)\footnote{\url{https://www.cs.cornell.edu/people/pabo/movie-review-data/}}~\citep{pang-lee-2005-seeing}, split by \citet{DBLP:conf/kdd/TangQM15}, is a binary sentiment analysis dataset on sentence level (mean sequence length and SD: $25 \pm 11$). 
Table~\ref{tab:datasets} shows the dataset characteristics.
\begin{table}[ht]
\small
    \centering
    \caption{Characteristics of text classification datasets}\label{tab:datasets}
    \begin{tabular}{lrrrr}
    \toprule
    Dataset & N       & \#Train & \#Test  & \#Classes \\
    \midrule                                            
    20ng    & 18,846  & 11,314  & 7,532   & 20        \\
    R8      & 7,674   & 5,485   & 2,189   & 8         \\
    R52     & 9,100   & 6,532   & 2,568   & 52        \\
    ohsumed & 7,400   & 3,357   & 4,043   & 23        \\
    MR      & 10,662  & 7,108   & 3,554   & 2         \\
    \bottomrule
    \end{tabular}
\end{table}

\subsection{Inductive and Transductive Setups}
We distinguish between a transductive and an inductive setup 
for text categorization.
In the transductive setup, as used in TextGCN, the test documents are visible and actually used for the preprocessing step. 
In the inductive setting, the test documents remain unseen until test time (\ie they are not available for preprocessing). 
We report the scores of the graph-based models for both setups from the literature, where available.
BoW-based and sequence-based models are inherently inductive.
\citet{DBLP:conf/wsdm/RageshSIBL21} have evaluated a variant of TextGCN that is capable of inductive learning, which we include in our results, too. 

\subsection{Procedure and Hyperparameter Settings}

We have extracted accuracy scores from the literature according to our systematic selection from \Secref{sec:sota}. 
Below, we provide a detailed description of the procedure for the models that we have run ourselves.
We borrow the tokenization strategy from BERT~\cite{DBLP:conf/naacl/DevlinCLT19} along with its uncased vocabulary. 
The tokenizer relies primarily on
WordPiece~\cite{wordpiece} for a high coverage while maintaining a small vocabulary.

\textit{Training our BoW-Models.}~
Our \mlp has one hidden layer with 1,024 rectified linear units (one input-to-hidden and one hidden-to-output layer). We apply dropout after each hidden layer, notably also after the initial embedding layer. Only for GloVe+\mlp, neither dropout nor ReLU is applied to the \emph{frozen} pretrained embeddings but only on subsequent layers.
The variant \mlp-2 has two ReLU-activated hidden layers (three layers in total) with $1,024$ hidden units each.
While this might be overparameterized for single-label text classification tasks with few classes, we rely on recent findings that overparameterization leads to better generalization~\cite{DBLP:journals/corr/abs-1805-12076,DBLP:conf/iclr/NakkiranKBYBS20}. 
In pre-experiments, we realized that MLPs are not very sensitive to hyperparameter choices.
Therefore, we optimize cross-entropy with Adam~\cite{DBLP:journals/corr/KingmaB14} and its default learning rate of $10^{-3}$, a linearly decaying learning rate schedule and train for a high amount of steps \cite{DBLP:conf/iclr/NakkiranKBYBS20} (we use 100 epochs) with small batch sizes (we use 16) for sufficient stochasticity, along with a dropout ratio of $0.5$.

\textit{Fine-tuning our BERT models.}
For BERT and DistilBERT, we fine-tune for 10 epochs with a linearly decaying learning rate of $5 \cdot 10^{-5}$ and an effective batch size of 128 via gradient accumulation of 8 x 16 batches. We truncate all inputs to 512 tokens.
To isolate the influence of word order on BERT's performance, we conduct two further ablations.
First, we set all position embeddings to zero and disable their gradient (\textbf{BERT w/o pos ids}). By doing this, we force BERT to operate on a bag-of-words without any notion of word order or position.
Second, we shuffle each sequence to augment the training data. We use this augmentation strategy to increase the number of training examples by a factor of two (\textbf{BERT w/ shuf. augm.}).

\subsection{Measures}
We report accuracy as evaluation metric, which is equivalent to Micro-F1 in single-label classification (see Appendix~\ref{app:microf1-equals-accuracy}). 
We repeat all experiments five times with different random initialization of the parameters and report the mean and standard deviation of these five runs.

\section{Results}\label{sec:results}

\subsection{Effectiveness}

Table~\ref{tab:results_textclf_micro} shows the accuracy scores for the text categorization models on the five datasets.
All graph-based models in the transductive setting show similar accuracy scores (maximum difference is 2 points). As expected, the scores decrease in the inductive setting up to a point where they are matched or even outperformed by our \mlp.

In the inductive setting, the WideMLP models perform best among the BoW models, in particular, TFIDF+WideMLP and WideMLP on an unweighted BoW.
The best-performing graph-based model is HyperGAT, yet DADGNN has a slight advantage on R8, R52, and MR.
For the sequence-based models, BERT attains the highest scores, closely followed by DistilBERT.

\begin{table*}[ht]
    \small
    \centering
    \caption{Accuracy and standard deviation on text classification datasets. Column ``Provenance'' reports the source.}\label{tab:results_textclf_micro}
    \begin{tabular}{llllllr}

  \toprule
  \textbf{Inductive Setting}  & \textbf{20ng} & \textbf{R8} & \textbf{R52} & \textbf{ohsumed} & \textbf{MR} & \textbf{Provenance}\\
    \midrule

\textit{BoW-Models} & & & & & & \\
    Log. Regression & 83.70 & 93.33 & 90.65 & 61.14 & 76.28 & \citet{DBLP:conf/wsdm/RageshSIBL21}\\
    SWEM & 85.16 (0.29) & 95.32 (0.26) & 92.94 (0.24) &  63.12 (0.55) & 76.65 (0.63) & \citet{DBLP:conf/emnlp/DingWLLL20}\\
    fastText & 79.38 (0.30) & 96.13 (0.21) &  92.81 (0.09) &  57.70 (0.49) & 75.14 (0.20) & \citet{DBLP:conf/emnlp/DingWLLL20}\\

     TF-IDF + \mlp & 84.20 (0.16) & 97.08 (0.16) & 93.67 (0.23) & 66.06 (0.29) & 76.32 (0.17)& own experiment\\
    \mlp  & 83.31 (0.22) & 97.27 (0.12) & 93.89 (0.16) & 63.95 (0.13) & 76.72 (0.26) & own experiment\\ 
    \mlp-2  & 81.02 (0.23) & 96.61 (1.22) & 93.98 (0.23) & 61.71 (0.33) & 75.91 (0.51) & own experiment\\
    GloVe+\mlp  & 76.80 (0.11) & 96.44 (0.08) & 93.58 (0.06) & 61.36 (0.22) & 75.96 (0.17) & own experiment\\ 
    GloVe+\mlp-2  & 76.33 (0.18) & 96.50 (0.14) & 93.19 (0.11) & 61.65 (0.27) & 75.72 (0.45)& own experiment\\

\midrule
    \textit{Graph-based Models} & & & & & & \\
    
         TextGCN & 80.88 (0.54) & 94.00 (0.40) & 89.39 (0.38) & 56.32 (1.36) & 74.60 (0.43) & \citet{DBLP:conf/wsdm/RageshSIBL21}\\
     HeteGCN & 84.59 (0.14) & 97.17 (0.33) & 93.89 (0.45) & 63.79 (0.80) & 75.62 (0.26) & \citet{DBLP:conf/wsdm/RageshSIBL21}\\
     HyperGAT & 86.62 (0.16) & 97.07 (0.23) & 94.98 (0.27) & 69.90 (0.34) & 78.32 (0.27) &\citet{DBLP:conf/wsdm/RageshSIBL21}\\
    DADGNN & --- & 98.15 (0.16) & 95.16 (0.22) & --- & 78.64 (0.29) & \citet{DBLP:conf/emnlp/LiuGG0F21}\\
     \midrule
     
    \textit{Seq.-based Models} & & & & & & \\
     
    LSTM (pretrain) & 75.43 (1.72) & 96.09 (0.19) &  90.48 (0.86) &  51.10 (1.50) & 77.33 (0.89) & \citet{DBLP:conf/emnlp/DingWLLL20}\\
    
    DistilBERT  & 86.24 (0.26) & 97.89 (0.15) & 95.34 (0.08) & 69.08 (0.60) & 85.10 (0.33)& own experiment\\
    BERT & 87.21 (0.18) & 98.03 (0.24) & 96.17 (0.33) & 71.46 (0.54) & 86.61 (0.38)& own experiment\\
    BERT w/o pos emb  & 81.47 (0.49) & 97.39 (0.20) & 94.70 (0.27) & 65.18 (1.53) & 80.35 (0.20)& own experiment \\
    BERT w/ shuf. augm.  & 86.46 (0.42) & 98.07 (0.21) & 96.48 (0.18) & 70.94 (0.60) & 86.23 (0.33) & own experiment\\
    
        \toprule
     \textbf{Transductive Setting} & \textbf{20ng} & \textbf{R8} & \textbf{R52} & \textbf{ohsumed} & \textbf{MR} & \textbf{Provenance}\\
  \midrule
      \textit{Graph-based Models} & & & & & & \\

    TextGCN& 86.34 & 97.07 & 93.56 & 68.36 & 76.74 & \citet{DBLP:conf/aaai/YaoM019}\\
    SGC& 88.5 (0.1)& 97.2 (0.1) & 94.0 (0.2) & 68.5 (0.3) & 75.9 (0.3) & \citet{DBLP:conf/icml/WuSZFYW19}\\
    TensorGCN& 87.74 & 98.04  & 95.05  & 70.11  & 77.91 & \citet{DBLP:conf/aaai/LiuYZWL20}\\
    HeteGCN& 87.15 (0.15) & 97.24 (0.51) & 94.35 (0.25) & 68.11 (0.70) & 76.71 (0.33) & \citet{DBLP:conf/wsdm/RageshSIBL21}\\
    
    \bottomrule
    \end{tabular}
\end{table*}

The strong performance of \mlp rivals all graph-based techniques reported in the literature, in particular, the recently published graph-inducing methods.
MLP only falls behind HyperGAT, which relies on topic models to set up the graph. 
Another observation is that $1$ hidden layer (but wide) is sufficient for the tasks, as the scores for MLP variants with $2$ hidden layers are lower.
We further observe that both pure BoW and TF-IDF weighted BoW lead to better results than approaches that exploit pretrained word embeddings such as GloVe-MLP, fastText, and SWEM.

With its immense pretraining, BERT yields the overall highest scores, closely followed by DistilBERT. DistilBERT outperforms HyperGAT by 7 points on the MR dataset while being on par on the others.
BERT outperforms the strongest graph-based competitor, HyperGAT, by 8 points on MR, 1.5 points on ohsumed, 1 point on R52 and R8, and 0.5 points on 20ng.

Our results further confirm that position embeddings are important for BERT with a notable decrease when those are omitted. Augmenting the data with shuffled sequences has led to neither a consistent decrease nor increase in performance.

\subsection{Efficiency}

\paragraph{Parameter Count  of the Models}
Table~\ref{tab:num_params} lists the parameter counts of the models.
Even though the MLP is fully-connected on top of a bag-of-words with the dimensionality of the vocabulary size, it has only half of the parameters as DistilBERT and a quarter of the parameters of BERT. Using TF-IDF does not change the number of model parameters.
Due to the high vocabulary size, GloVe-based models have a high number of parameters, but the majority of those is frozen, \ie does not get gradient updates during training.

\begin{table}[ht]
    \small
    \centering
    \caption{Parameter counts of the models}\label{tab:num_params}
    \begin{tabular}{lr}
    \toprule
    Model & \#parameters  \\
    \midrule
         \mlp & 31.3M\\
         \mlp-2 & 32.3M \\
         GloVe+\mlp & 575,2M (frozen) + 0.3M\\
         GloVe+\mlp-2 & 575,2M (frozen) +  1.3M\\
         DistilBERT & 66M\\
         BERT & 110M\\
     \bottomrule
    \end{tabular}
\end{table}

\paragraph{Runtime Performance of the Models}
We provide the \emph{total} running times in Table~\ref{tab:runtime} as observed while conducting the experiments on a single NVIDIA A100-SXM4-40GB card. 
All \mlp variants are an order of magnitude faster than DistilBERT when considering the average runtime \emph{per epoch}. DistilBERT is twice as fast as the original BERT. The transformers are only faster than BoW models on the MR dataset. This is because the sequences in the MR dataset are much shorter and less $\mathcal{O}(L^2)$ attention weights have to be computed.

\begin{table*}[ht]
    \small
    \centering
    \caption{Total runtime (training+inference). Average of five runs rounded to minutes.}\label{tab:runtime} \begin{tabular}{llccccc}
    \toprule
    Model & \#epochs & 20ng & R8 & R52 & ohsumed & MR  \\
    \midrule
         \mlp & 100 &  7min & 3min & 4min & 3min & 4min\\
         TF-IDF+\mlp & 100 & 9min & 4min & 4min & 3min & 4min\\
         \mlp-2 & 100 &  9min & 5min & 5min & 3min & 6min\\
         GloVe+\mlp & 100 &  6min & 3min & 4min & 3min & 4min\\
         GloVe+\mlp-2  & 100 & 6min & 4min & 4min & 3min & 4min\\
         DistilBERT & 10 & 8min & 4min & 5min & 3min & 1min \\
         BERT & 10 & 15min & 7min & 8min & 5min & 2min\\
     \bottomrule
    \end{tabular}
\end{table*}

\section{Discussion}\label{sec:discussion}

\paragraph{Key Insights}
Our experiments show that our MLP models using BoW outperform the recent graph-based models TextGCN and HeteGCN in an inductive text classification setting.
Furthermore, the MLP models are comparable to HyperGAT. 
Only transformer-based BERT and DistilBERT models outperform our MLP and set a new state-of-the-art.
This result is important for two reasons:
First, the strong performance of a pure BoW-MLP questions the added value of synthetic graphs in models like TextGCN to the text categorization task.
Only HyperGAT, which uses the expensive Latent Dirichlet Allocation for computing the graph, slightly outperforms our BoW-\mlp in two out of five datasets.
Thus, we argue that using strong baseline models for text classification is important to assess the true scientific advancement~\cite{DBLP:conf/recsys/DacremaCJ19}. 

Second, in contrast to conventional wisdom~\cite{DBLP:conf/acl/IyyerMBD15}, we find that pretrained embeddings, \eg GloVe, can have a detrimental effect when compared to using an MLP with one wide hidden layer. Such an MLP circumvents the bottleneck of the small dimensionality of word embeddings and has a higher capacity.
Furthermore, we experiment with more hidden layers (see \mlp-2), but do not observe any improvement when the single hidden layer is sufficiently wide. A possible explanation is that already a single hidden layer is sufficient to approximate any compact function to an arbitrary degree of accuracy depending on the width of the hidden layer~\cite{DBLP:journals/mcss/Cybenko89}.

Finally, a new state-of-the-art is set by the transformer model BERT, which is not very surprising.
However, as our efficiency analysis shows, the MLPs require only a fraction of the parameters and are faster in their combined training and inference time except for the MR dataset. The attention mechanism of (standard) Transformers is quadratic in the sequence length, which leads to slower processing of long sequences. With larger batches, the speed of the MLP could be increased even further.

\paragraph{Detailed Discussion of Results}
Graph-based models come with high training costs, as not only the graph has to be first computed, but also a GNN has to be trained. For standard GNN methods, the whole graph has to fit into the GPU memory and mini-batching is nontrivial, but possible with dedicated sampling techniques for GNNs~\cite{DBLP:conf/icml/FeyLWL21}.
Furthermore, the original TextGCN is inherently transductive, \ie it has to be retrained whenever new documents appear. Strictly transductive models are effectively useless in practice~\cite{DBLP:journals/corr/abs-1906-12330} except for applications, in which a partially labeled corpus needs to be fully annotated.
However, recent extensions such as HeteGCN, HyperGAT, and DADGNN already relax this constraint and enable inductive learning. Nevertheless, word-document graphs require $\mathcal{O}(N^2)$ space, where $N$ is the number of documents plus the vocabulary size, which is a hurdle for large-scale applications.

There are also tasks where the \emph{natural} structure of the graph data provides more information than the mere text, \eg citations networks or connections in social graphs. 
In such cases, the performance of graph neural networks is the state of the art~\cite{DBLP:conf/iclr/KipfW17,velickovic2018graph} and are superior to MLPs that use only the node features and not the graph structure~\cite{DBLP:journals/corr/abs-1811-05868}. 
GNNs also find application in various NLP tasks, other than classification~\cite{gnn-for-nlp-survey}.

An interesting factor is the ability of the models to capture word order.
BoW models disregard word order entirely and yield good results, but still fall behind order-aware Transformer models.
In an extensive study, \citet{DBLP:conf/acl/BaroniBLKC18} have shown that memorizing the word content (which words appear at all) is most indicative of downstream task performance.
\citet{sinha2021masked} have experimented with pretraining BERT by disabling word order during pretraining and show that it makes surprisingly little difference for fine-tuning. In their study, word order is preserved during fine-tuning. In our experiments, we have conducted complementary experiments: we have used a BERT model that is pretrained with word order, but we have deactivated the position encoding during fine-tuning.
Our results show that there is a notable drop in performance but the model does not fail completely.

Other NLP tasks such as question answering~\cite{DBLP:conf/emnlp/RajpurkarZLL16} or natural language inference~\cite{DBLP:conf/iclr/WangSMHLB19} can also be regarded as text classification on a technical level. 
Here, the positional information of the sequence is more important than for pure topical text categorization.
One can expect that BoW-based models perform worse than sequence-based models. 

\paragraph{Generalizability}
We expect that similar observations would be made on other text classification datasets because we have already covered a range of different characteristics: long and short texts, topical categorization (20ng, Reuters, and Ohsumed) and sentiment prediction (MR) in the domains of forum postings, news, movie reviews, and medical abstracts.
Our results are in line with those from other fields, who have reported a resurgence of MLPs.
For example, in business prediction, an MLP baseline outperforms various other Deep Learning models~\cite{IJCNN-Ishwar-2021,yedida2021simple}.
In computer vision, \citet{DBLP:journals/corr/abs-2105-01601} and \citet{DBLP:journals/corr/abs-2105-02723} proposed attention-free MLP models that are on par with the Vision Transformer~\citep{visiontransformer}.
In natural language processing, \citet{DBLP:journals/corr/abs-2105-08050} show similar results, while acknowledging that a small attention module is necessary for some tasks.

\paragraph{Threats to Validity}
We acknowledge that the experimental datasets are limited to English.
While word order is important in the English language, 
it is notable that methods that discard word order still work well for text categorization.
Another possible bias is the comparability of the results. However, we carefully checked all relevant parameters such as the train/test split, the number of classes in the datasets, if datasets have been pre-processed in such a way that, e.\,g., makes a task easier like reducing the number of classes, the training procedure, and the reported evaluation metrics.
Regarding our efficency analysis, we made sure to report numbers for the parameter count \emph{and} a measure for the speed other than FLOPs, as recommended by \citet{efficiency-misnomer}. Since runtime is heavily dependant on training parameters such as batch size, we complement this with asymptotic complexity.

\paragraph{Practical Impact and Future Work}
Our study has an immediate impact on practitioners who seek to employ robust text categorization models in research projects and in industrial operational environments. 
Furthermore, we advocate to use an MLP baseline in future text categorization research, for which we provide concrete guidelines in Appendix~\ref{app:guidelines}.
As future work, it would be interesting to analyze multi-label classification tasks and to compare with hierarchical text categorization methods 
~\cite{DBLP:conf/www/PengLHLBWS018,DBLP:conf/www/XiaoLS19}. 
Another interesting yet challenging setting would be few-shot classification~\cite{DBLP:conf/nips/BrownMRSKDNSSAA20}.

\section{Conclusion}
We argue that a wide multi-layer perceptron enhanced with today's best practices should be considered as a strong baseline for text classification tasks. In fact, the experiments show that our \mlp is oftentimes on-par or even better than recently proposed models that synthesize a graph structure from the text.
 
The source code is available online:
\url{https://github.com/lgalke/text-clf-baselines}

\section*{Ethical Considerations}
The focus of this work is text classification.
Potential risks that apply to text classification in general also apply to this work.
Nonetheless, we present alternatives to commonly used pretrained language models,
which suffer from various sources of bias due to the large and poorly manageable data used for pretraining~\cite{BenderEtAl_2021_DangersStochasticParrots}.
In contrast, the presented alternatives render full control over the training data and, thus, contribute to circumvent the biases otherwise introduced during pretraining.  

\bibliography{main}
\bibliographystyle{acl_natbib}

\newpage
\appendix

\section{Practical Guidelines for Designing a WideMLP}\label{app:guidelines}
On the basis of our results, we provide recommendations for designing a \mlp baseline.

\paragraph{Tokenization}
We recommend using modern subword tokenizers such as BERT-like WordPiece or SentencePiece that yield a high coverage while needing a relatively small vocabulary.

\paragraph{Input Representation}
In contrast to conventional wisdom~\cite{DBLP:conf/acl/IyyerMBD15}, we find that pretrained embeddings, \eg GloVe, can have a detrimental effect when compared to using an MLP with one wide hidden layer. Such an MLP circumvents the bottleneck of the small dimensionality of word embeddings and has a higher capacity.

\paragraph{Depth vs.\@ Width}
In text classification, width seems more important than depth. We recommend to use a single, wide hidden layer, \ie one input-to-hidden and one hidden-to-output layer, \eg with 1,024 hidden units and ReLU activation. 
While this might be overparameterized for single-label text classification tasks with few classes, we rely on recent findings that overparameterization leads to better generalization~\cite{DBLP:journals/corr/abs-1805-12076,DBLP:conf/iclr/NakkiranKBYBS20}. 

We further motivate the choice of using wide layers with results from multi-label text classification~\cite{DBLP:conf/kcap/GalkeMSBS17}, which has shown that a (wide)
MLP outperforms all tested classical baselines such as SVMs, k-Nearest Neighbors, and logistic regression. Follow-up work~\cite{DBLP:conf/jcdl/MaiGS18} then found that also CNN and LSTM do not substantially improve over the wide MLP.

Having a fully-connected layer on-top of a bag-of-words leads to a high number of learnable parameters. Still, the wide first input-to-hidden layer can be implemented efficiently by using an embedding layer followed by aggregation, which avoids large matrix multiplications. 

In our experiments, we did not observe any improvement with more hidden layers (\mlp-2), as suggested by \citet{DBLP:conf/acl/IyyerMBD15}, but it might be beneficial for other, more challenging datasets.

\paragraph{Optimization and Regularization}
We seek to find an optimization strategy that does not require dataset-specific hyperparameter tuning. 
This comprises optimizing cross-entropy with Adam~\cite{DBLP:journals/corr/KingmaB14} and default learning rate $10^{-3}$, a linearly decaying learning rate schedule and training for a high amount of steps \cite{DBLP:conf/iclr/NakkiranKBYBS20} (we use 100 epochs) with small batch sizes (we use 16) for sufficient stochasticity.

For regularization during this prolonged training, we suggest to use a high dropout ratio of $0.5$. 
Regarding initialization, we rely on framework defaults, \ie $\mathcal{N}(0,1)$ for the initial embedding layer and random uniform $\mathcal{U}(-\sqrt{d_\mathrm{input}}, \sqrt{d_\mathrm{output}})$ for subsequent layers' weight and bias parameters.

\section{Connection between BoW-MLP and TextGCN}\label{app:textgcn-vs-mlp}
TextGCN uses the PMI matrix to set up edge weights for word-word connections.
A single layer Text-GCN is a BoW-MLP, except for the document embedding. The one-hop neighbors are words which are aggregated after a nonlinear transform. The basic GCN equation $\mH = \sigma (\hat\mA \mX \mW)$ reveals that the order of transformation and neighborhood aggregation is irrelevant.
The document embedding implies that TextGCN is a semisupervised technique.
Truly new documents, as in inductive learning scenarios, would need a special treatment such as using an all zero embedding vector. 

A two-layer MLP can be characterized by the equation
$\hat{y} = \mW^{(2)} \sigma (\mW^{(1)} \vx + \vb^{(1)}) + \vb^{(2)}$.
On bag-of-words inputs, the first layer $\mW^{(1)} \vx + \vb^{(1)}$ can be replaced by an equivalent embedding layer with weighting (\eg TF-IDF or length normalization) being applied during aggregation of the embedding vectors. 

The first layer of TextGCN is equivalent to aggregating embedding vectors.
A standard GCN layer with shared weights has the form (assuming self-loops have been inserted)

\[
\vh_i =  \sum_{j \in N(i)} a_{ij} \mW^{(1)} \vx_j + \vb^{(1)}
\]
Now in TextGCN node features are given by the identity, such that $\vx_j$ is one-hot. Then we can rewrite the first layer of Text-GCN as an aggregation of embeddings $\mE$. We gain 
\[
\vh_i = \sum_{j \in N(i)} a_{ij} \mE_j
\]

as  $\mW \vx + \vb$ may again be replaced by an embedding matrix if applied to one hot vectors $\vx$.
Now $\mE$ contains two types of embedding vectors: word embeddings and document embeddings corresponding to word nodes and document nodes. We see that the first layer of TextGCN is essentially an aggregation of word embeddings plus the document embedding.
Only with a second layer, TextGCN considers the embedding of other documents whose words are connected to the present documents' words.

\section{Equivalence of Micro-F1 and Accuracy in Multiclass Classification}\label{app:microf1-equals-accuracy}
\newcommand{\tp}{\mathrm{TP}}
\newcommand{\tn}{\mathrm{TN}}
\newcommand{\fp}{\mathrm{FP}}
\newcommand{\fn}{\mathrm{FN}}
In multiclass classification, we have a single true label for each instance and the predictions are constrained to a single prediction per instance. As a consequence, the measures accuracy and Micro-F1 coincide to the same formula.

Micro-F1 aggregates true positives (TP), true negatives (TN), false positives (FP), and false negatives (FN) globally.
It can be expressed as:

\[
\mathrm{Micro\text{-}F_1} = \frac{2 \sum_c \tp_c}{2 \sum_c \tp_c + \sum_c \fp_c + \sum_c \fn_c},
\]
where $c$ iterates over all classes.

While the accuracy can be expressed as:

\[
\mathrm{Acc} = \frac{\sum_c \tp_c + \sum_c \tn_c}{\sum_c \tp_c + \sum_c \tn_c + \sum_c \fp_c + \sum_c \fn_c}
\]

In multiclass classification, every true positive is also a true negative for all other classes.
When summing those up over the entire dataset, we obtain
\[
    \sum_c \tp_c = \sum_c \tn_c.
\]

Thus, we can rewrite 
\[
2 \sum_c \tp_c = \sum_c \tp_c + \sum_c\tn_c
\]
and see that the Micro-F1 and accuracy are equivalent in the multiclass (a.k.a. single-label) case.

\end{document}